\documentclass[10pt]{article}
\usepackage{graphicx}  
\usepackage[svgnames]{xcolor}
\usepackage{bbm}
\usepackage[authoryear]{natbib}
\usepackage{graphicx}
\usepackage{algorithm}
\usepackage{algpseudocode}  
\usepackage{appendix}
\usepackage[utf8]{inputenc}
\usepackage{amsmath}
\usepackage{amsfonts}
\usepackage{amssymb}
\usepackage{mathtools}  
\usepackage[version=4]{mhchem}
\usepackage{stmaryrd}
\usepackage{bbold}
\usepackage{enumitem}
\setlist[itemize]{label=\textbullet}
\usepackage{subcaption}
\usepackage{authblk}    
\usepackage[margin=0.8in]{geometry}
\usepackage{hyperref}
\usepackage{longtable}
\usepackage{booktabs}
\usepackage{multirow}
\usepackage{url}
\usepackage{setspace}  
\begin{document}
\title{Data Mixing as Mixture Experiment:\\Response Surface Methodology and Optimal Design\\for Large Language Model Pretraining}
\author[1]{Yicheng Mao}
  \author[2,*]{Hongru Du}
  \affil[1]{Department of Mathematics and Statistics, University of Calgary, University Drive NW, Calgary, T2N 1N4, Canada}
  \affil[2]{Department of Systems and Information Engineering, University of Virginia, Charlottesville, VA, USA}
  \affil[*]{Correspondence: hongrudu@virginia.edu}

\date{}
\maketitle
\doublespacing 
\begin{abstract}
Data mixing is a central design problem in large language model pretraining: given a fixed token budget, practitioners must decide how much data to allocate to each domain. Recent proxy-based methods address this problem by training small models on candidate mixtures, fitting a response model, and using the response to select mixtures for larger-scale training. We show that this workflow has the structure of a classical mixture experiment. Under this view, data domains are mixture components, token shares are component proportions, proxy-training runs are experimental design points, and validation loss defines a response surface over the probability simplex. We develop this formulation using sparse second-order Scheff\'{e} response-surface models and construct model-robust $\mathcal{I}$-optimal designs for proxy data-mixing experiments. Using RegMix as an empirical case study, we demonstrate how the framework can both interpret observed mixture responses and design more efficient proxy experiments. The Scheff\'{e} analysis shows that domain value is strongly relational: several domains that are weak under additive effects become favourable through pairwise interactions, especially through combinations with web-derived text. The sparse Scheff\'{e} model preserves mixture rankings across model scales and remains competitive with a flexible machine-learning predictor while providing an explicit decomposition of additive and interaction effects. In a simulation study calibrated to observed proxy-training responses, model-robust $\mathcal{I}$-optimal designs recover the relevant mixture ordering after removing about 25\% of the original proxy runs. These results suggest that LLM data mixing should be treated not only as a prediction problem, but also as an experimental-design problem in which the proxy mixtures themselves can be chosen to improve statistical efficiency.
\end{abstract}


\section{Introduction}
\label{sec:intro}

The performance of a large language model depends heavily on the composition of its pretraining corpus. Modern pretraining datasets are assembled from multiple data domains such as web text, code, scientific articles, and books, and the proportion allocated to each domain affects both validation loss and downstream task performance \citep{gao2021pile, longpre2023flan}. Because the total token budget is often fixed, increasing the share of one domain reduces the share available to the others. The choice of data mixture is therefore a constrained allocation problem, and it has become a central design question in LLM pretraining. At the scale of modern pretraining runs, suboptimal mixture choices translate directly into wasted compute and degraded model capability, making principled mixture selection a problem of both scientific and practical importance.

Recent work has begun to treat data mixture weights not as fixed corpus-construction choices, but as quantities that can be estimated, adapted, or optimised during the training pipeline. For example, DoReMi \citep{xie2023doremi} uses a small proxy model and group distributionally robust optimisation to learn domain weights that are then used for larger-scale pretraining. More recent work makes the mixture-performance relationship even more explicit by modelling how mixture proportions affect language-model performance. Data Mixing Laws \citep{ye2024datamixinglaws} fit predictive functions that relate sampled mixtures to language modelling loss, while BiMix \citep{ge2024bimix} extends this idea by modelling the joint effect of mixture proportions and data volume. RegMix \citep{liu2025regmix} similarly trains small proxy models on sampled mixtures and fits regression models to predict the performance of unseen mixtures at larger scales. Taken together, these studies show that data mixing is increasingly being treated as a measurable allocation problem: mixture proportions are varied, model responses are observed or predicted, and the resulting information is used to guide future mixture selection.

The key observation of this paper is that this allocation problem has the geometry of a classical mixture experiment.  
The predictors in data mixing are not ordinary covariates. They are proportions that must be non-negative and sum to one, so all feasible mixtures lie on a probability simplex. The same geometry appears in classical mixture experiments, where the response depends on the relative proportions of several components rather than on independently adjustable inputs \citep{cornell2002experiments}. The classical examples are formulation problems: the taste of a food product depends on the proportions of its ingredients, and the performance of a chemical product depends on the proportions of its constituent compounds \citep{Goos2019choice,Furlanetto2011Mixture}. The same mixture-experiment logic has also been used for more abstract allocation problems, where the ``components'' are not physical ingredients but shares of a fixed resource. For example, advertising media-mix studies allocate a fixed advertising budget across media channels to quantify channel effects and cross-media synergy \citep{goos2019advertising}, while mobility-budget studies allocate a fixed travel budget across alternative transport modes \citep{zijlstra2019mixture}. In all these cases, the effect of one component cannot be interpreted independently of the others because increasing one share necessarily reduces the remaining shares. LLM data mixing follows the same logic: the components are data domains, the mixture proportions are token shares, and the response is validation loss or downstream performance after training.

This mixture-experiment interpretation has two methodological consequences. First, it changes how the response surface can be specified. Since the predictors are constrained proportions, the fitted model should respect the simplex geometry rather than treating domain shares as ordinary unconstrained covariates. This motivates the use of Scheff\'{e} response-surface models, which are defined directly on the simplex \citep{scheffe1958}. In the present context, they provide a way to decompose the fitted loss surface into additive domain contributions and pairwise departures from additivity. This is important because a domain may be weak as an additive contributor but useful when combined with another domain. Flexible machine-learning predictors may capture such interactions implicitly, but they do not usually express them as interpretable quantities.

Second, the mixture-experiment interpretation changes how proxy mixtures can be selected. In many proxy-based data-mixing workflows, candidate mixtures are sampled randomly, for example, from Dirichlet distributions.
This is convenient and scalable, but it does not ask whether those mixtures are the most informative design points for estimating a response surface. If proxy runs are experimental design points rather than merely random samples, then their locations on the simplex can be chosen to improve prediction over the mixture region. Since the practical goal is to predict the performance of unseen mixtures, $\mathcal{I}$-optimality is a natural criterion because it targets low average prediction variance over the experimental region \citep{goos2011optimal}. Moreover, because the appropriate response-surface order is not known before the proxy experiment is run, a model-robust design can be used to balance performance across additive and interaction-inclusive candidate models \citep{yu2008model}.

This paper develops a mixture-experiment framework for LLM data mixing, in which data sources are components, token shares are mixture proportions, proxy-training runs are design points, and model performance is the response surface. 
Using the publicly available RegMix proxy-training data as an empirical case study, we fit sparse second-order Scheff\'{e} models to estimate interpretable additive and interaction effects, evaluate whether these structured response surfaces preserve mixture rankings across model scales, and construct model-robust $\mathcal{I}$-optimal designs to test whether useful mixture orderings can be recovered with fewer proxy runs.  
Although the empirical analysis is grounded in the RegMix setting, the methodological framework applies to any proxy-based data-mixing workflow in which a fixed training budget is allocated across data sources.


The contribution of the paper is threefold. First, it formalises LLM data mixing as a mixture response-surface experiment on the probability simplex. This connects proxy-based data-mixing workflows to the statistical literature on mixture modelling and optimal experimental design. Second, it shows how sparse Scheff\'{e} models can expose additive domain effects and pairwise domain interactions that are otherwise hidden inside black-box predictors. Third, it demonstrates that optimal design can be used to improve the efficiency of proxy data-mixing experiments, reducing the proxy-training budget needed to recover reliable mixture rankings and suggesting that how proxy mixtures are chosen deserves as much attention as how the resulting data are modelled.

The remainder of the paper is organised as follows. Section~\ref{sec:regmix-mixture} formalises proxy-based data mixing as a mixture response-surface experiment. Section~\ref{sec:methods} introduces the sparse Scheff\'{e} response-surface model and the model-robust $\mathcal{I}$-optimal design criterion. Section~\ref{sec:results-regmix-response-surface} reports the Scheff\'{e} re-analysis of the RegMix data and its cross-scale rank-preservation performance. Section~\ref{sec:results-design} evaluates model-robust $\mathcal{I}$-optimal designs for proxy data-mixing experiments. Section~\ref{sec:discussion} concludes with limitations and directions for future work.

\section{Proxy Data Mixing as a Mixture Experiment}
\label{sec:regmix-mixture}

Consider a fixed pretraining token budget distributed across $K$ data domains. Let
$$
\mathbf{x}=(x_1,\ldots,x_K)^\top
$$
denote the vector of domain proportions. Each candidate mixture satisfies
$$
x_i \geq 0,\qquad \sum_{i=1}^{K}x_i=1.
$$
Thus, the space of feasible data mixtures is the $(K-1)$-dimensional probability simplex. The practical goal is to identify a mixture $\mathbf{x}$ that optimises a specified training objective, such as validation loss or downstream task performance, subject to this fixed-budget simplex constraint.

A proxy-based data-mixing study can be viewed as an experiment on this simplex. A set of candidate mixtures is selected, a small model is trained or evaluated at each mixture, and the resulting validation loss or downstream score is recorded as the response. A predictive model is then fitted from mixture proportions to model performance and used to guide the selection of future mixtures for larger-scale training. Under this view, data sources are mixture components, token shares are component proportions, proxy-training runs are experimental design points, and the fitted predictor is a response surface over the simplex.

The RegMix proxy-training data \citep{liu2025regmix} provide a useful empirical instance of this general structure. RegMix samples candidate mixtures from Dirichlet distributions, trains a small proxy LLM on each sampled mixture, records the target value to be optimised, and fits a regression model using mixture proportions as features and target values as labels. The fitted model is then used to predict the performance of a large simulated pool of candidate mixtures, from which the best predicted mixture is selected for larger-scale training. In the main experiment analysed here, RegMix uses 17 available domains from the Pile dataset \citep{gao2021pile}, trains 512 proxy models with 1M non-embedding parameters on 1B tokens each, and evaluates the fitted regression model on held-out mixtures at 1M, 60M, and 1B model scales. The corresponding domain labels used in our analysis are listed in Table~\ref{tab:domains}. All empirical analyses in this paper are conducted within this specific experimental setting; how far the empirical findings generalise to other domain taxonomies, corpora, or training objectives is an open question we return to in the Discussion.

\begin{table}[h!]
\centering
\caption{Pile domain labels used in the RegMix main experiment.}
\label{tab:domains}
\small
\begin{tabular}{@{}lll@{}}
\toprule
Label used in this paper & Pile component & Brief description \\
\midrule
\texttt{arxiv} & ArXiv & Scientific papers and preprints \\
\texttt{freelaw} & FreeLaw & Legal opinions and court documents \\
\texttt{nih\_exporter} & NIH ExPorter & NIH grant and project records \\
\texttt{pubmed\_central} & PubMed Central & Full-text biomedical articles \\
\texttt{wikipedia\_en} & Wikipedia (en) & English Wikipedia articles \\
\texttt{dm\_mathematics} & DM Mathematics & Mathematical problem and proof text \\
\texttt{github} & GitHub & Source code and code-related text \\
\texttt{philpapers} & PhilPapers & Philosophy papers and bibliographic text \\
\texttt{stackexchange} & Stack Exchange & Question-answer forum content \\
\texttt{enron\_emails} & Enron Emails & Corporate email correspondence \\
\texttt{gutenberg\_pg\_19} & Gutenberg (PG-19) & Public-domain books \\
\texttt{pile\_cc} & Pile-CC & CommonCrawl-derived web text \\
\texttt{ubuntu\_irc} & Ubuntu IRC & Online chat logs from Ubuntu IRC \\
\texttt{europarl} & EuroParl & European Parliament proceedings \\
\texttt{hackernews} & HackerNews & Technology-oriented discussion forum text \\
\texttt{pubmed\_abstracts} & PubMed Abstracts & Biomedical article abstracts \\
\texttt{uspto\_backgrounds} & USPTO Backgrounds & Patent background sections \\
\bottomrule
\end{tabular}
\end{table}

The proxy-to-large-scale transfer used in this setting relies on rank preservation across scales. In practical terms, a mixture that performs relatively well for a small model trained under a proxy budget is expected to remain relatively strong for a larger model trained under a larger budget. This assumption makes it possible to use many cheaper proxy runs to guide expensive large-scale mixture selection. It also explains why RegMix evaluates its regression models primarily through rank correlation across scales, rather than only through absolute loss prediction \citep{liu2025regmix}.

The mixture-experiment interpretation makes the structure of such proxy workflows explicit. The sampled proxy mixtures are design points, the validation losses are responses, and the fitted model is a response surface over the simplex. The predictors are not ordinary unconstrained covariates: increasing one domain proportion necessarily reduces the total proportion available to the others. This compositional constraint is precisely the defining feature of a classical mixture experiment. Table~\ref{tab:mapping} summarises this correspondence.

\begin{table}[h!]
\centering
\caption{Correspondence between proxy-based data mixing and classical mixture experiment methodology.}
\label{tab:mapping}
\small
\begin{tabular}{@{}ll@{}}
\toprule
Proxy-based data-mixing workflow & Mixture experiment interpretation \\
\midrule
Data domains & Mixture components \\
Domain token shares $x_i$ & Component proportions \\
Constraint $\sum_i x_i=1$ & Simplex constraint \\
Sampled proxy-training mixtures & Experimental design points \\
Proxy-model validation loss or downstream score & Continuous response \\
Predictive model fitted to mixture--performance pairs & Response-surface model \\
Candidate-mixture search & Response-surface optimisation \\
Random or heuristic mixture sampling & Experimental design strategy \\
\bottomrule
\end{tabular}
\end{table}

This reinterpretation matters for two reasons. First, it changes how the response surface can be modelled. Standard regression or machine-learning predictors can be useful for ranking mixtures, but they do not necessarily express the compositional structure of the predictors. For example, RegMix compares linear regression and LightGBM \citep{ke2017lightgbm}. A linear model is simple but additive in its basic form, while LightGBM can capture nonlinearities and interactions only implicitly through the fitted tree ensemble. Scheff\'{e} response-surface models provide a natural alternative: they are defined directly on the simplex and can represent pairwise domain interactions through explicit coefficients. This is useful when the value of a data source depends not only on its own share, but also on the other sources with which it is combined.

Second, the mixture-experiment view changes how proxy mixtures can be selected. Random or heuristic sampling is convenient and scalable, but it does not ask whether the selected mixtures are the most informative design points for estimating a response surface. If proxy runs are experimental design points rather than merely sampled candidates, then their locations on the simplex can be chosen to improve prediction over the mixture region. In particular, $\mathcal{I}$-optimality targets low average prediction variance over a specified mixture region, making it a natural design criterion when the aim is to predict the performance of unseen mixtures. If such designs can achieve comparable ranking performance with fewer proxy runs, they could reduce the token and compute cost of data-mixing experiments.

These observations motivate the empirical analysis that follows. We use the RegMix proxy-training data as a public case study for applying mixture response-surface modelling and optimal design to LLM data mixing. Specifically, we fit sparse quadratic Scheff\'{e} models to obtain an interpretable response surface, evaluate whether this model preserves cross-scale mixture rankings, and examine whether $\mathcal{I}$-optimal mixture designs can reduce the number of proxy runs required to recover prediction-relevant mixture orderings. The purpose is not to reformulate RegMix as a method, but to use its proxy-training data to illustrate a broader statistical framework for fixed-budget data mixing.

\section{Methods}
\label{sec:methods}

\subsection{Sparse Scheff\'{e} response-surface modelling}
\label{sec:methods-scheffe}

We model proxy-based data-mixing experiments as response-surface problems on the mixture simplex.
Let $\mathbf{x}=(x_1,\ldots,x_K)^\top$ denote a data mixture over $K$ domains, with $x_i\geq 0$ and $\sum_i x_i=1$. The experimental region is therefore the $(K-1)$-dimensional simplex. Under this constraint, the usual intercept is not separately identifiable from the component proportions, since the constant term can be written as $1=\sum_i x_i$. We therefore use the canonical Scheff\'{e} response-surface models for mixture experiments, which omit the ordinary intercept \citep{scheffe1958}.

The first-order Scheff\'{e} model is
\begin{equation}
    E(Y\mid \mathbf{x})=\sum_{i=1}^{K}\beta_i x_i,
    \label{eq:scheffe-first-order}
\end{equation}
where $Y$ denotes the validation loss of a proxy LLM trained on mixture $\mathbf{x}$. In this parameterisation, $\beta_i$ corresponds to the fitted response at the pure-domain vertex for domain $i$. This interpretation is model-based, since practical pretraining mixtures need not include pure-domain configurations.

To represent pairwise domain interactions, we use the second-order Scheff\'{e} model,
\begin{equation}
E(Y\mid \mathbf{x})
=
\sum_{i=1}^{K}\beta_i x_i
+
\sum_{i=1}^{K-1}\sum_{j=i+1}^{K}
\beta_{ij}x_i x_j.
\label{eq:scheffe-second-order}
\end{equation}
The coefficient $\beta_{ij}$ measures the departure from additivity when domains $i$ and $j$ are combined. Since the response is validation loss, $\beta_{ij}<0$ indicates that the pair lowers the fitted loss relative to the additive expectation, whereas $\beta_{ij}>0$ indicates that the pair raises it. This interpretation is conditional on the chosen validation objective.

The collective contribution of pairwise interactions can be assessed by comparing the first-order model in Eq.~\eqref{eq:scheffe-first-order} with the full second-order model in Eq.~\eqref{eq:scheffe-second-order}. We use the nested-model statistic
\begin{equation}
T_F
=
\frac{
(\mathrm{RSS}_{\mathrm{lin}}-\mathrm{RSS}_{\mathrm{quad}})
/
(p_{\mathrm{quad}}-p_{\mathrm{lin}})
}{
\mathrm{RSS}_{\mathrm{quad}}
/
(n-p_{\mathrm{quad}})
}.
\label{eq:nested-f-test}
\end{equation}
Here $\mathrm{RSS}_{\mathrm{lin}}$ and $\mathrm{RSS}_{\mathrm{quad}}$ are the residual sums of squares for the first-order and full second-order models, while $p_{\mathrm{lin}}$ and $p_{\mathrm{quad}}$ are the corresponding numbers of fitted terms. Eq.~\eqref{eq:nested-f-test} evaluates whether the pairwise interaction terms improve the response surface as a group. It is not used as a significance test for individual interaction coefficients.

In the empirical RegMix case study, $K=17$, so the full second-order model contains $17+\binom{17}{2}=153$ terms. Even when interactions are useful as a group, interpreting all pairwise coefficients is undesirable because many terms may be weak, redundant, or unstable. We therefore use an $L_1$-penalised second-order Scheff\'{e} model to obtain a sparse response surface:
\begin{equation}
\hat{\boldsymbol{\beta}}
=
\arg\min_{\boldsymbol{\beta}}
\left\{
\|\mathbf{y}-\mathbf{Z}\boldsymbol{\beta}\|_2^2
+
\lambda\|\boldsymbol{\beta}\|_1
\right\},
\label{eq:lasso-scheffe}
\end{equation}
where $\mathbf{y}$ is the vector of validation losses, $\mathbf{Z}$ is the full second-order Scheff\'{e} model matrix, and $\lambda$ controls the strength of the sparsity penalty. The $L_1$ penalty shrinks weak coefficients to zero, producing a smaller set of selected domain main effects and pairwise interactions \citep{tibshirani1996regression}. The resulting sparse model is used to interpret the fitted response surface.

\subsection{Model-robust \texorpdfstring{$\mathcal{I}$}{I}-optimal mixture design}
\label{sec:methods-design}

In mixture experiments, optimal design criteria are commonly used to select informative experimental configurations. Two standard criteria are $\mathcal{D}$-optimality and $\mathcal{I}$-optimality. The former targets precise estimation of model parameters, while the latter targets accurate prediction over the experimental region \citep{cornell2002experiments,goos2011optimal}. Since proxy-based data-mixing studies use pilot experiments to predict the performance of unseen data mixtures, this paper focuses on $\mathcal{I}$-optimality. A detailed description of $\mathcal{D}$-optimality is provided in the Supplementary Information.

In the present setting, a proxy-training design is a collection of mixture configurations used to train small proxy LLMs and estimate the response surface. For a proxy-training design $\mathbf{X}=(\mathbf{x}_1,\ldots,\mathbf{x}_n)^\top\in\mathbb{R}^{n\times K}$, let $\mathbf{f}(\mathbf{x})$ denote the response-surface feature vector for a candidate mixture $\mathbf{x}$, and let $\mathbf{Z}(\mathbf{X})$ denote the corresponding model matrix.
$\mathcal{I}$-optimal designs minimise the average prediction variance over the experimental region. For a linear response-surface model with homoscedastic errors, this criterion can be written as
\begin{equation}
\mathcal{I}(\mathbf{X})
=
\log
\left\{
\operatorname{tr}
\left[
\mathbf{M}(\mathbf{X})^{-1}\mathbf{W}
\right]
\right\},
\label{eq:i-criterion}
\end{equation}
where $\mathbf{M}(\mathbf{X})$ is the information matrix and $\mathbf{W}$ is the moment matrix over the simplex. The information matrix is
\begin{equation}
\mathbf{M}(\mathbf{X})
=
\mathbf{Z}(\mathbf{X})^\top \mathbf{Z}(\mathbf{X}).
\label{eq:information-matrix}
\end{equation}
The moment matrix is defined as
\begin{equation}
\mathbf{W}
=
\int_{\mathcal{X}}
\mathbf{f}(\mathbf{x})\mathbf{f}(\mathbf{x})^\top
\,d\mathbf{x},
\label{eq:moment-matrix}
\end{equation}
where $\mathcal{X}$ is the feasible mixture region. Each element of $\mathbf{W}$ is therefore an integral of a monomial over the simplex. For non-negative integer powers $a_1,\ldots,a_K$, these moments are given by
\begin{equation}
\int_{\mathcal{X}}
x_1^{a_1}x_2^{a_2}\cdots x_K^{a_K}
\,d\mathbf{x}
=
\frac{
\prod_{i=1}^{K}\Gamma(a_i+1)
}{
\Gamma\left(K+\sum_{i=1}^{K}a_i\right)
}.
\label{eq:simplex-moments}
\end{equation}
To illustrate the construction of $\mathbf{W}$, consider a second-order Scheff\'{e} model with three mixture components. The feature vector is
$$
\mathbf{f}(\mathbf{x})
=
(x_1,x_2,x_3,x_1x_2,x_1x_3,x_2x_3)^\top .
$$
Using Eq.~\eqref{eq:simplex-moments}, the corresponding moment matrix is
$$
\mathbf{W}
=
\begin{pmatrix}
\frac{1}{12} & \frac{1}{24} & \frac{1}{24} & \frac{1}{60} & \frac{1}{60} & \frac{1}{120} \\
\frac{1}{24} & \frac{1}{12} & \frac{1}{24} & \frac{1}{60} & \frac{1}{120} & \frac{1}{60} \\
\frac{1}{24} & \frac{1}{24} & \frac{1}{12} & \frac{1}{120} & \frac{1}{60} & \frac{1}{60} \\
\frac{1}{60} & \frac{1}{60} & \frac{1}{120} & \frac{1}{180} & \frac{1}{360} & \frac{1}{360} \\
\frac{1}{60} & \frac{1}{120} & \frac{1}{60} & \frac{1}{360} & \frac{1}{180} & \frac{1}{360} \\
\frac{1}{120} & \frac{1}{60} & \frac{1}{60} & \frac{1}{360} & \frac{1}{360} & \frac{1}{180}
\end{pmatrix}.
$$
Lower values of $\mathcal{I}(\mathbf{X})$ indicate lower average prediction variance over the mixture region.

To compare a design $\mathbf{X}$ with a reference design $\mathbf{X}^{*}$ under the same response-surface model, we use relative $\mathcal{I}$-efficiency,
\begin{equation}
\mathcal{I}\text{-eff}(\mathbf{X},\mathbf{X}^{*})
=
\exp
\left[
\mathcal{I}(\mathbf{X}^{*})-\mathcal{I}(\mathbf{X})
\right].
\label{eq:i-efficiency}
\end{equation}
Values greater than one indicate that $\mathbf{X}$ has lower average prediction variance than the reference design.

During the design stage, information about domain interactions is limited, making it difficult to determine whether a first-order or second-order response surface is more appropriate. Since a design optimised for one candidate model can be inefficient under another model form \citep{Atkinson1996}, we adopt a model-robust $\mathcal{I}$-optimal design strategy \citep{yu2008model,mao2026partial}. The criterion is constructed over two candidate Scheff\'{e} models: the first-order model and the full second-order model.

Let $\mathcal{I}_1(\mathbf{X})$ and $\mathcal{I}_2(\mathbf{X})$ denote the $\mathcal{I}$-criteria computed under the first-order and second-order candidate models, respectively. The model-robust criterion is defined as
\begin{equation}
\mathcal{I}_{\mathrm{rob}}(\mathbf{X})
=
\frac{1}{2}\mathcal{I}_1(\mathbf{X})
+
\frac{1}{2}\mathcal{I}_2(\mathbf{X}).
\label{eq:i-robust}
\end{equation}
This criterion favours proxy-training designs that retain good average prediction precision under both additive and interaction-inclusive response surfaces.

To construct the model-robust $\mathcal{I}$-optimal designs, we use the simulated annealing algorithm for optimal mixture designs proposed by \citet{Mao2025mixture}. This algorithm is suitable for the present simplex-constrained optimisation problem and has been shown to produce high-quality mixture designs relative to conventional hill-climbing search strategies. Detailed pseudocode and implementation details are provided in the Supplementary Information.


\section{Empirical Response-Surface Analysis}
\label{sec:results-regmix-response-surface}

This section illustrates the Scheff\'{e} mixture-response-surface formulation using the RegMix proxy-training data.
The goal is to decompose the fitted response surface into additive domain contributions and pairwise departures from additivity, and then assess whether this structured surface preserves mixture rankings across model scales.

\subsection{Scheff\'{e} re-analysis of the response surface}
\label{sec:results-scheffe-interpretation}

The first-order terms of the Scheff\'{e} model summarise the additive contribution of each domain to the fitted mixture response surface. Because the response is validation loss, smaller first-order coefficients correspond to lower fitted loss under the additive component. These coefficients are defined on the simplex and should not be read as unconstrained marginal effects or as estimates of pure-domain training performance.

\begin{figure}[h!]
\centering
\includegraphics[width=\textwidth]{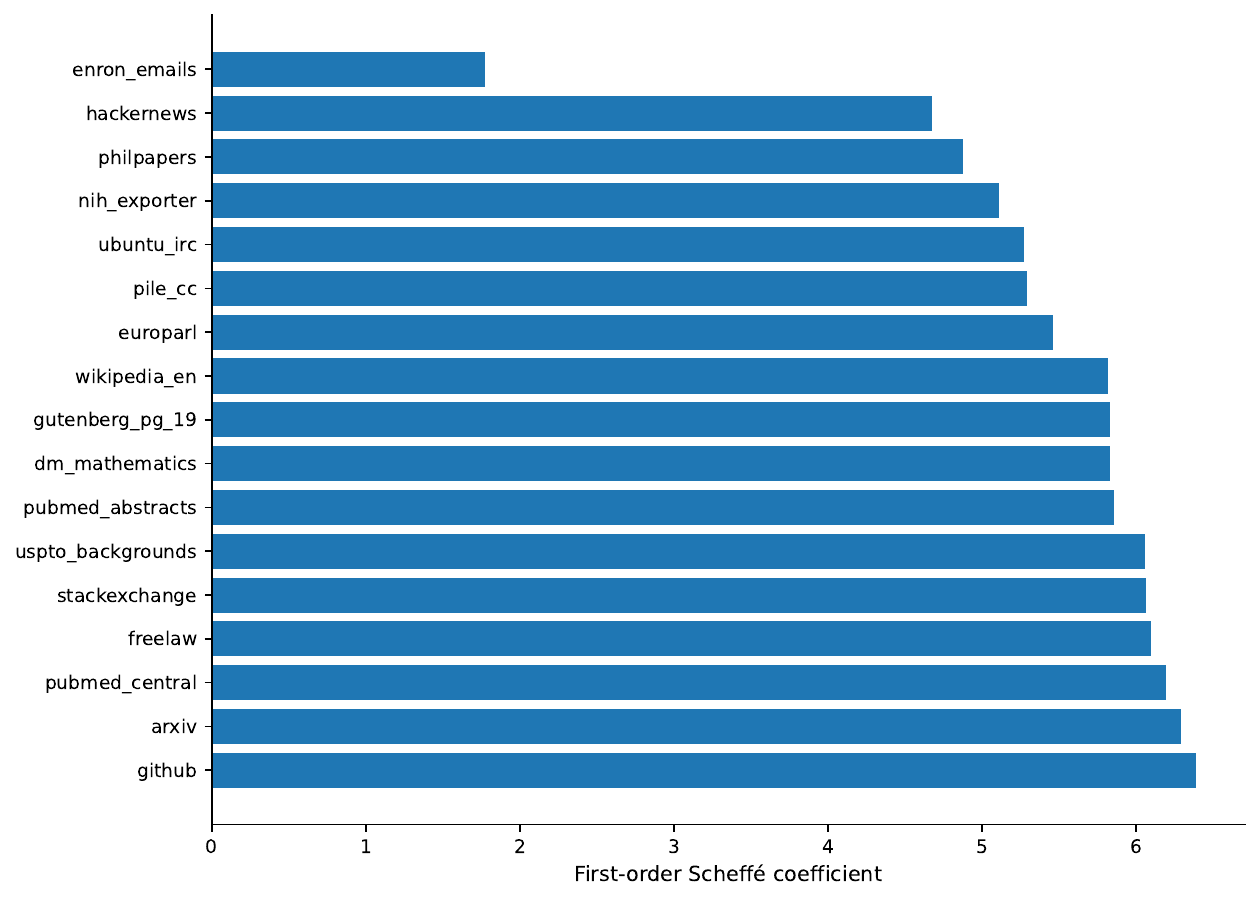}
\caption{First-order domain coefficients from the sparse Scheff\'{e} response-surface model. Because the response is validation loss, smaller coefficients correspond to lower fitted loss under the additive component.}
\label{fig:scheffe-main-effects}
\end{figure}

Figure~\ref{fig:scheffe-main-effects} reports the first-order coefficients from the sparse Scheff\'{e} model. The additive ordering departs from conventional data-quality assumptions. The smallest coefficient belongs to \texttt{enron\_emails}, followed by \texttt{hackernews}, \texttt{philpapers}, \texttt{nih\_exporter}, \texttt{ubuntu\_irc}, and \texttt{pile\_cc}. Several substantively rich or technical domains, including \texttt{github}, \texttt{arxiv}, \texttt{pubmed\_central}, \texttt{freelaw}, and \texttt{stackexchange}, carry larger first-order coefficients, and are therefore not the most favourable additive contributors to reducing the target validation loss.

The additive surface is informative but incomplete. Comparing the first-order and full second-order Scheff\'{e} models yields $F = 6.5145$ on $136$ interaction and $359$ residual degrees of freedom ($p < 0.001$), so the pairwise terms improve the fitted surface beyond the first-order Scheff\'{e} model. 

The sparse second-order model was fitted using the $L_1$ penalty in Eq.~\eqref{eq:lasso-scheffe}, with $\lambda$ selected by 10-fold cross-validation using mean squared prediction error over a logarithmic grid from $10^{-5}$ to $1$.
The selected value was $\lambda=1.4175\times 10^{-5}$. At this penalty level, the model retained 80 terms, including all 17 first-order terms and 63 pairwise interactions. Among these interactions, 51 have negative coefficients and 12 have positive coefficients. Negative coefficients indicate fitted loss reductions relative to the additive expectation, while positive coefficients indicate fitted loss increases.

\begin{figure}[h!]
\centering
\includegraphics[width=\textwidth]{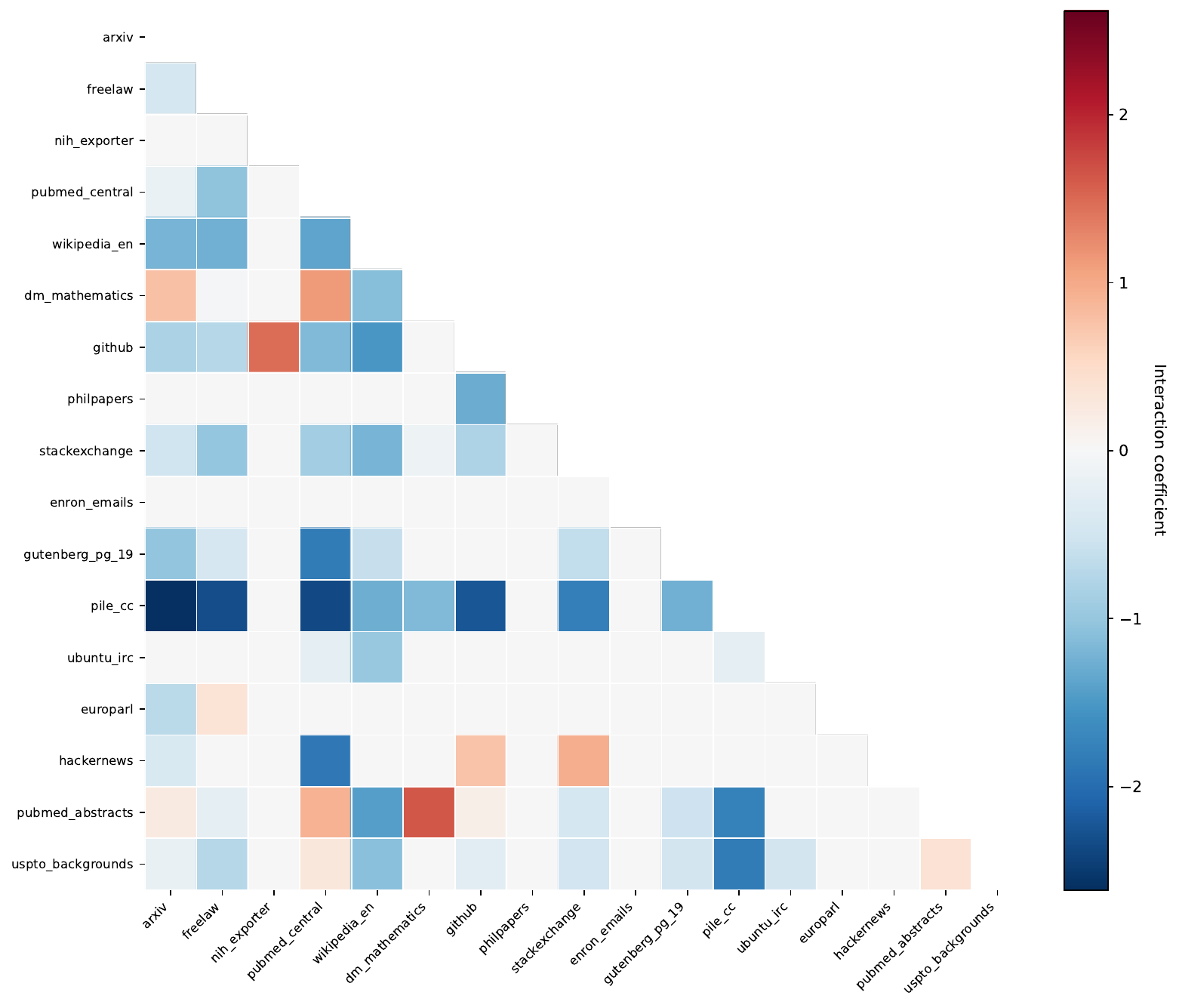}
\caption{Selected pairwise interaction coefficients from the sparse second-order Scheff\'{e} model. Blue cells denote negative coefficients, corresponding to lower fitted validation loss relative to the additive mixture baseline. Red cells denote positive coefficients, corresponding to higher fitted validation loss relative to the additive baseline. White or near-white cells indicate interactions that are weak or not selected by the sparse model.}
\label{fig:scheffe-interaction-heatmap}
\end{figure}

Figure~\ref{fig:scheffe-interaction-heatmap} shows that the interaction surface is sparse and structured, dominated by a few pronounced effects centered on \texttt{pile\_cc} rather than by a diffuse field of weak pairwise terms. Its strongest negative interactions, in descending magnitude, are with \texttt{arxiv} ($\beta = -2.62$), \texttt{pubmed\_central} ($-2.36$), \texttt{freelaw} ($-2.31$), \texttt{github} ($-2.24$), \texttt{uspto\_backgrounds} ($-1.83$), \texttt{stackexchange} ($-1.78$), and \texttt{pubmed\_abstracts} ($-1.78$), with further negative terms involving \texttt{wikipedia\_en}, \texttt{gutenberg\_pg\_19}, and \texttt{dm\_mathematics}; the four largest coefficients in the entire model are all \texttt{pile\_cc} pairings. Several of these partner domains carry large first-order coefficients, so their contribution is better explained by complementarity with \texttt{pile\_cc} than by additive effects.

This contrast between main effects and interactions is the central result. Domains such as \texttt{arxiv}, \texttt{pubmed\_central}, \texttt{freelaw}, and \texttt{github} are unfavourable in the additive component yet become favourable once combined with \texttt{pile\_cc}. The positive interactions are fewer, smaller, and more localised. The three largest are \texttt{dm\_mathematics} with \texttt{pubmed\_abstracts} ($\beta = +1.62$), \texttt{nih\_exporter} with \texttt{github} ($+1.45$), and \texttt{pubmed\_central} with \texttt{dm\_mathematics} ($+1.11$). A positive coefficient means that the two domains together yield higher fitted loss than their additive contributions predict, so allocating weight to both at once is less efficient than the additive surface suggests. The likely cause is redundancy or competition under the target objective: the two domains supply overlapping signal, so weight spent on both returns less than their separate contributions imply.

This decomposition illustrates what is gained by treating proxy data mixing as a mixture response-surface experiment. Flexible predictors can capture nonlinearities and interactions inside the fitted function, but they do not necessarily expose those interactions as interpretable quantities. In the present case study, the Scheff\'{e} decomposition represents the interaction structure as a set of signed pairwise coefficients, making it possible to identify which domains are complementary, which compete, and by how much. The strongest such structure is the role of \texttt{pile\_cc}: it carries only a moderate first-order coefficient yet enters the four largest interactions in the model, all negative. Its value is therefore relational rather than standalone. This helps explain why web-derived text can be highly useful in this dataset: its advantage operates largely through complementarity with other domains rather than through a dominant additive effect.

The same decomposition also changes how individual domains should be interpreted. A domain that appears weak under the additive component may still be valuable when combined with other domains, while a domain that appears favourable in isolation may contribute less through interactions. For example, \texttt{philpapers} has a favourable first-order coefficient but enters only one strong interaction, whereas domains such as \texttt{arxiv} and \texttt{pubmed\_central} look weak additively but become more favourable once their interactions with \texttt{pile\_cc} are included. The relevant quantity is therefore not whether a domain is good or bad in isolation, but how its additive and interaction components combine within the simplex. The Scheff\'{e} surface makes both components directly inspectable.

\subsection{Rank preservation across model scales}
\label{sec:results-rank-preservation}

We evaluate cross-scale transfer by comparing predicted and observed mixture orderings at the 1M, 60M, and 1B evaluation scales. Let $\hat{y}_{ig}$ denote the predicted response for mixture $i$ under fitted response-surface model $g$, and let $y_i^{(s)}$ denote the observed response for the same mixture at scale $s\in\{\mathrm{1M},\mathrm{60M},\mathrm{1B}\}$. We report Spearman's rank correlation,
\begin{equation}
\rho_s(g)
=
\operatorname{corr}_{\mathrm{Spearman}}
\left(
\{\hat{y}_{ig}\}_{i=1}^{n_s},
\{y_i^{(s)}\}_{i=1}^{n_s}
\right),
\label{eq:spearman-rank-preservation}
\end{equation}
and pairwise ranking accuracy,
\begin{equation}
\mathrm{PRA}_s(g)
=
\frac{2}{n_s(n_s-1)}
\sum_{1\leq i<j\leq n_s}
\mathbf{1}
\left[
\operatorname{sign}\{y_i^{(s)}-y_j^{(s)}\}
=
\operatorname{sign}\{\hat{y}_{ig}-\hat{y}_{jg}\}
\right].
\label{eq:pairwise-ranking-accuracy}
\end{equation}
Spearman's $\rho_s(g)$ measures agreement between the overall predicted and observed rankings. $\mathrm{PRA}_s(g)$ measures the proportion of mixture pairs whose relative ordering is correctly predicted. All three models are fitted on the 512 proxy mixtures at the 1M scale and evaluated on held-out mixtures at each scale. The number of evaluation mixtures is 256 at the 1M and 60M scales and 64 at the 1B scale.

To quantify uncertainty in these rank-preservation metrics, we report 95\% percentile bootstrap confidence intervals based on 2,000 bootstrap replicates. For Spearman's $\rho$, bootstrap replicates resample evaluation mixtures with replacement and recompute the rank correlation using the fixed fitted model predictions. For $\mathrm{PRA}$, bootstrap replicates resample the binary correctness indicators over unordered mixture pairs with replacement, using the mixture pair as the resampling unit.  

\begin{table}[h!]
\centering
\caption{Cross-scale rank-preservation performance of response-surface models fitted to the RegMix proxy-training data. Values in brackets are 95\% percentile bootstrap confidence intervals. The intervals are defined by the 2.5th and 97.5th percentiles of the resulting bootstrap distributions.}
\label{tab:model-comparison-rank}
\small
\begin{tabular}{@{}lcccc@{}}
\toprule
Model & Scale & $n$ & $\rho$ & PRA \\
\midrule
First-order Scheff\'{e}        & 1M  & 256 & 0.902 [0.869, 0.925] & 0.867 [0.863, 0.870] \\
First-order Scheff\'{e}        & 60M & 256 & 0.892 [0.859, 0.917] & 0.859 [0.855, 0.863] \\
First-order Scheff\'{e}        & 1B  & 64  & 0.879 [0.767, 0.942] & 0.864 [0.849, 0.878] \\
\addlinespace
Sparse Second-order Scheff\'{e} & 1M  & 256 & 0.937 [0.914, 0.952] & 0.894 [0.891, 0.898] \\
Sparse Second-order Scheff\'{e} & 60M & 256 & 0.939 [0.916, 0.954] & 0.896 [0.893, 0.900] \\
Sparse Second-order Scheff\'{e} & 1B  & 64  & 0.975 [0.949, 0.985] & 0.937 [0.926, 0.948] \\
\addlinespace
LightGBM                       & 1M  & 256 & 0.990 [0.986, 0.992] & 0.959 [0.957, 0.961] \\
LightGBM                       & 60M & 256 & 0.986 [0.979, 0.989] & 0.951 [0.949, 0.953] \\
LightGBM                       & 1B  & 64  & 0.962 [0.913, 0.982] & 0.927 [0.916, 0.938] \\
\bottomrule
\end{tabular}
\end{table}

Table~\ref{tab:model-comparison-rank} compares the first-order Scheff\'{e} model, the sparse second-order Scheff\'{e} model, and LightGBM under both metrics. The first-order model already recovers most of the cross-scale ordering. Its Spearman correlations range from 0.879 to 0.902, and its PRA values remain above 0.85 at all three scales. This indicates that additive mixture structure captures a substantial part of the ranking signal. However, uncertainty is larger at the 1B scale, where the evaluation set contains only 64 mixtures. For the first-order model at 1B, for example, $\rho=0.879$ with a 95\% confidence interval of [0.767, 0.942].

Adding sparse second-order terms improves both metrics at all three scales. At 1M, Spearman's $\rho$ increases from 0.902 to 0.937, and PRA increases from 0.867 to 0.894. At 60M, the corresponding increases are from 0.892 to 0.939 for $\rho$ and from 0.859 to 0.896 for PRA. The largest point-estimate gain occurs at 1B, where $\rho$ increases from 0.879 to 0.975 and PRA increases from 0.864 to 0.937. The bootstrap intervals also support a consistent improvement of the sparse second-order model over the first-order model, particularly for PRA. These results suggest that the interaction structure identified in Section~\ref{sec:results-scheffe-interpretation} is relevant for cross-scale ranking, not only for interpreting the fitted response surface.

LightGBM performs best at the 1M and 60M scales, with $\rho=0.990$ and $\mathrm{PRA}=0.959$ at 1M, and $\rho=0.986$ and $\mathrm{PRA}=0.951$ at 60M. This is expected given its greater flexibility. At the 1B scale, the sparse second-order Scheff\'{e} model attains slightly higher point estimates than LightGBM, with $\rho=0.975$ and $\mathrm{PRA}=0.937$, compared with $\rho=0.962$ and $\mathrm{PRA}=0.927$ for LightGBM. However, this difference should be interpreted cautiously. The 1B evaluation set contains only 64 mixtures, and the bootstrap confidence intervals overlap for both metrics. The evidence therefore supports comparable performance at 1B, rather than a clear claim that the sparse Scheff\'{e} model outperforms LightGBM.

The main implication is that sparse Scheff\'{e} modelling provides a useful compromise between interpretability and cross-scale ranking performance. LightGBM remains the strongest flexible predictor at the 1M and 60M scales, while the sparse second-order Scheff\'{e} model remains competitive at 1B and provides an explicit decomposition of main effects and interactions. This makes the sparse Scheff\'{e} surface particularly useful when the goal is not only to predict promising mixtures, but also to understand which mixture components and interactions drive cross-scale transfer.

\section{Optimal Design for Proxy Data-Mixing Experiments}
\label{sec:results-design}

We next examine whether the proxy-training mixtures can be selected more efficiently than by random Dirichlet sampling. Using the simulated annealing algorithm described in Section~\ref{sec:methods-design}, we constructed model-robust $\mathcal{I}$-optimal designs for a sequence of design sizes $n \in 
\{160,192,224,256,288,320,352,384,416,448,480,512\}$. Each design was optimised under the optimality criterion introduced in Eq.\eqref{eq:i-robust}. We then compared these designs with the original 512-run proxy design from the empirical case study in terms of design geometry, relative $\mathcal{I}$-efficiency, and simulated ranking performance.

\begin{figure}[h!]
\centering
\includegraphics[width=\textwidth]{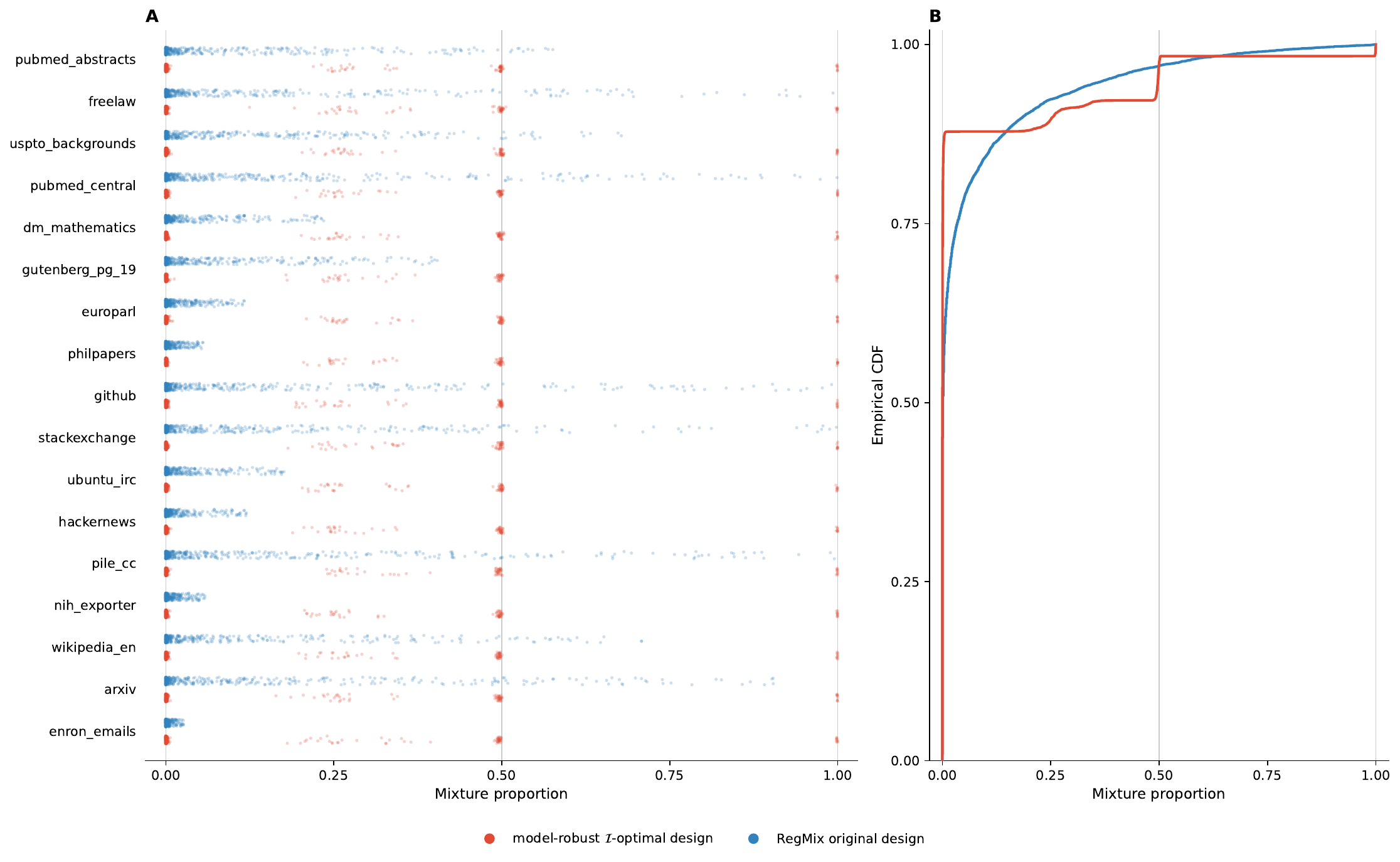}
\caption{Geometry of the model-robust $\mathcal{I}$-optimal design at $n=512$ compared with the original 512-run RegMix proxy design. Panel A shows the realised mixture proportions for each domain across all design points, with vertical jitter added for visibility. Panel B shows the empirical cumulative distribution function of all realised domain proportions pooled across domains and design points. The RegMix design spreads proportions more continuously over the simplex, whereas the model-robust $\mathcal{I}$-optimal design concentrates most proportions at the boundary values $0$, $0.5$, and $1$.}
\label{fig:design-spread}
\end{figure}

Figure~\ref{fig:design-spread} compares the geometry of the model-robust $\mathcal{I}$-optimal design at $n=512$ with the original 512-run reference design from the RegMix data. Panel A shows the realised proportions separately for each domain, while Panel B summarizes the marginal distribution of all realised domain proportions. The two designs have markedly different structures. The RegMix reference design spreads proportions continuously across the simplex, with many domains assigned small positive shares across runs. In contrast, the model-robust $\mathcal{I}$-optimal design is much more discrete: most realised domain proportions are placed at $0$, with additional support concentrated around $0.5$ and $1$. The empirical CDF in Panel B makes this contrast explicit, showing a sharp initial jump for the optimal design and step changes at the main support points. This boundary-concentrated structure is consistent with low-order mixture response-surface estimation, since simplex vertices identify first-order component effects and binary edge midpoints are informative for pairwise Scheff\'{e} interactions.

\begin{figure}[h!]
\centering
\includegraphics[width=0.75\textwidth]{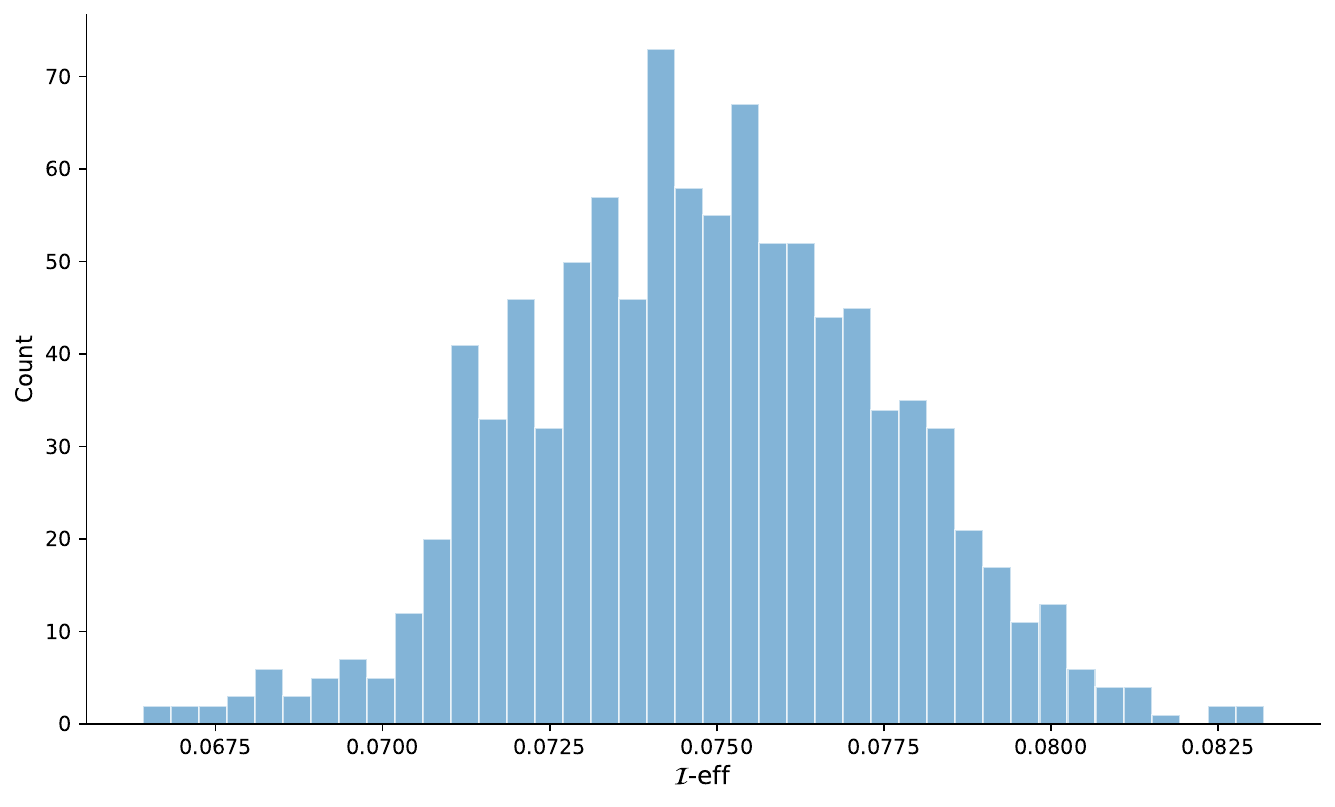}
\caption{Relative $\mathcal{I}$-efficiency of $1{,}000$ Dirichlet designs at $n=512$, evaluated under the sparse second-order Scheff\'{e} basis with the model-robust $\mathcal{I}$-optimal design as reference. Values below one indicate higher average prediction variance than the model-robust $\mathcal{I}$-optimal design.}
\label{fig:design-efficiency}
\end{figure}

We also compared the prediction-variance efficiency of the optimal design with 1,000 random Dirichlet designs of the same size. Let $\mathbf{X}^{\mathcal{I}}$ denote the model-robust $\mathcal{I}$-optimal design, and let $\mathbf{X}^{(b)}_{\mathrm{Dir}}$ denote the $b$th random Dirichlet design, for $b=1,\ldots,1000$. Following the relative $\mathcal{I}$-efficiency definition in Eq.~\eqref{eq:i-efficiency}, we computed
\[
\mathcal{I}\text{-eff}
\left(
\mathbf{X}^{(b)}_{\mathrm{Dir}},
\mathbf{X}^{\mathcal{I}}
\right)
=
\exp
\left[
\mathcal{I}(\mathbf{X}^{\mathcal{I}})
-
\mathcal{I}(\mathbf{X}^{(b)}_{\mathrm{Dir}})
\right],
\]
where both $\mathcal{I}$-criteria were evaluated under the sparse second-order Scheff\'{e} basis. Figure~\ref{fig:design-efficiency} shows that these relative efficiencies are concentrated around $0.07$ to $0.08$. Thus, under this basis, the random Dirichlet designs have substantially higher average prediction variance than the model-robust $\mathcal{I}$-optimal design.

\begin{figure}[h!]
\centering
\includegraphics[width=\textwidth]{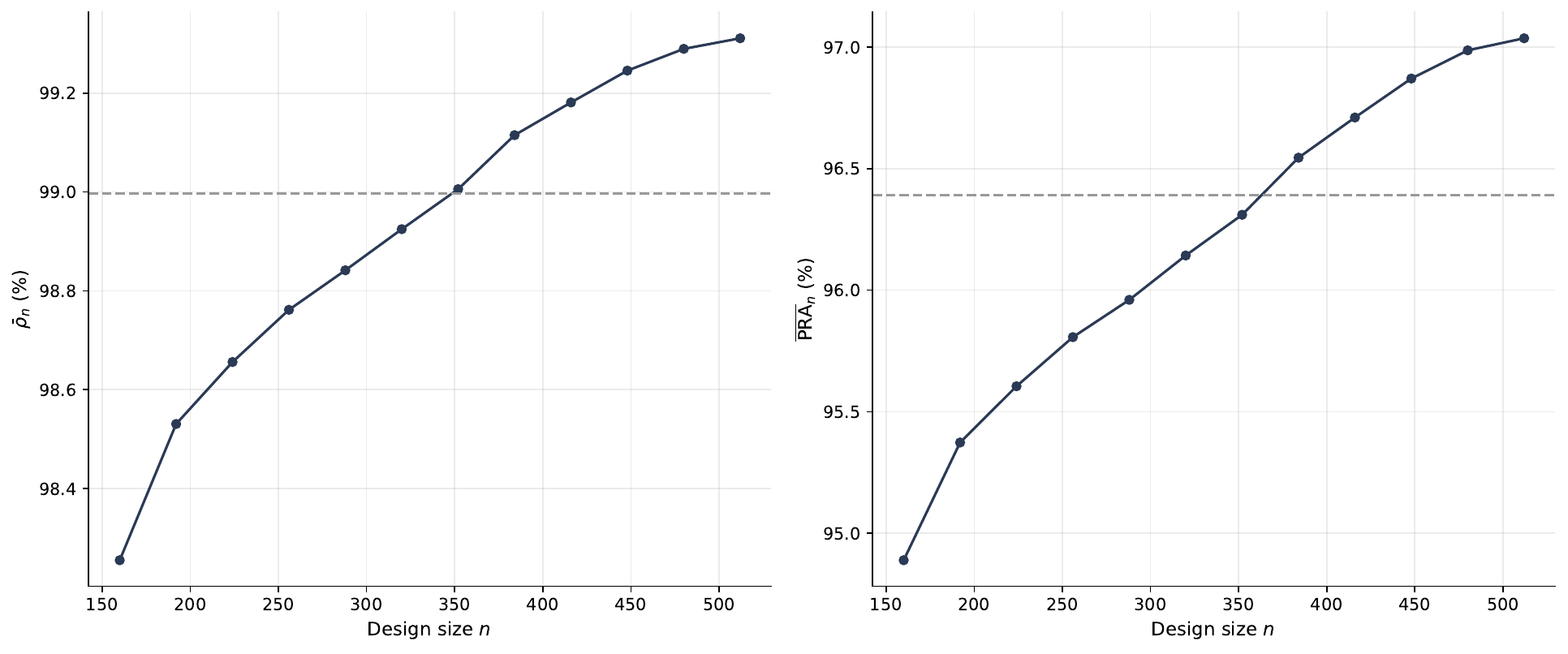}
\caption{Monte Carlo average ranking performance of the model-robust $\mathcal{I}$-optimal designs as a function of design size $n$ (solid line with markers), compared with the original RegMix design at $n=512$ (dashed line). Left: average Spearman rank correlation $\bar{\rho}_n$ between the predicted and reference 1B orderings. Right: average pairwise ranking accuracy $\overline{\mathrm{PRA}}_n$. Each point averages $1{,}000$ simulated datasets generated from the sparse Scheff\'{e} surface.}
\label{fig:design-performance}
\end{figure}

Finally, we evaluated whether this design advantage translates into mixture-ranking performance. We used the sparse second-order Scheff\'{e} surface fitted to the full RegMix proxy data as the data-generating surface. For each design size $n$, responses were generated at the corresponding model-robust $\mathcal{I}$-optimal design points with Gaussian noise estimated from the fitted residuals. The sparse Scheff\'{e} model was then refitted and used to rank the held-out 1B mixtures. This procedure was repeated for 1,000 simulated datasets at each design size. Let $\rho_n^{(b)}$ and $\mathrm{PRA}_n^{(b)}$ denote the Spearman rank correlation and pairwise ranking accuracy obtained from the $b$th simulated dataset, for $b=1,\ldots,1000$. We report the Monte Carlo averages
\[\bar{\rho}_n
=
\frac{1}{1000}
\sum_{b=1}^{1000}
\rho_n^{(b)},
\qquad
\overline{\mathrm{PRA}}_n
=
\frac{1}{1000}
\sum_{b=1}^{1000}
\mathrm{PRA}_n^{(b)}.\]
Figure~\ref{fig:design-performance} shows that both $\bar{\rho}_n$ and $\overline{\mathrm{PRA}}_n$ increase with design size. 
The model-robust $\mathcal{I}$-optimal designs reach and then exceed the original $n=512$ reference level with roughly 350 proxy runs, under both $\bar{\rho}_n$ and $\overline{\mathrm{PRA}}_n$.
At $n=512$, the model-robust $\mathcal{I}$-optimal design exceeds the RegMix reference under both metrics.

These results show that design-aware proxy mixture selection can reduce the proxy-training budget while preserving the relevant mixture ordering. In this simulation study, the model-robust $\mathcal{I}$-optimal designs match or exceed the original $n=512$ RegMix design after removing about 25\% of the proxy runs. Since each proxy run requires training a small model on a fixed token budget, this reduction translates directly into lower pilot-stage token and compute cost. The exact saving is specific to the fitted sparse Scheff\'{e} surface, but the result shows that optimal design can materially improve the efficiency of proxy-based data-mixing experiments.
As a sensitivity check, the Supplementary Material repeats the ranking evaluation under a first-order Scheff\'{e} data-generating surface and shows that the model-robust $\mathcal{I}$-optimal designs continue to outperform the original RegMix reference design.

\section{Discussion}
\label{sec:discussion}

This paper reframes LLM data mixing as a mixture experiment. Under this formulation, data domains are mixture components, proxy training runs are experimental design points, and validation loss defines a response surface over the probability simplex. This perspective turns data-mixing optimisation from a purely predictive task into a structured experimental-design problem. The empirical results show that this structure is useful in two ways: sparse Scheff\'{e} models provide an interpretable decomposition of domain effects and interactions, while model-robust $\mathcal{I}$-optimal designs can reduce the number of proxy runs needed to recover useful mixture rankings.

The empirical case study shows that domain value can be strongly relational. Several domains that are not favourable under the additive component become favourable through pairwise interactions, especially through combinations with \texttt{pile\_cc}. This suggests that pretraining data quality should not be understood only as a property of individual domains. A domain may be useful because of how it complements other domains under a fixed token budget. The Scheff\'{e} formulation makes this structure explicit by separating additive contributions from interaction effects, which are otherwise hidden inside flexible predictors such as LightGBM.

The design results indicate that the proxy stage itself can be made more efficient. Random Dirichlet sampling is convenient, but it is not necessarily the most informative way to estimate a mixture response surface. In the simulation-based case study, model-robust $\mathcal{I}$-optimal designs match or exceed the original RegMix reference after removing about 25\% of the proxy runs. This saving should not be read as universal, since it depends on the fitted sparse Scheff\'{e} surface and the assumed noise structure. The more general point is that mixture selection can benefit from design-aware sampling, especially when each proxy run carries a non-trivial token and compute cost.

Several limitations should be understood in light of the current stage of empirical research on LLM data mixing. Publicly available proxy-training datasets with systematically varied data mixtures remain limited, and generating new large-scale mixture experiments would require substantial token and compute resources. For this reason, the present empirical evaluation is based on one publicly available proxy-training dataset and a simulation-based comparison of alternative designs. The design evaluation uses the sparse Scheff\'{e} surface fitted to this dataset as the data-generating mechanism, so its conclusions depend on how well this fitted surface approximates the true training response. If higher-order interactions, threshold effects, or strong local irregularities dominate the response surface, a sparse quadratic Scheff\'{e} model may be too restrictive.
In addition, the optimal designs are constructed under an idealised mixture setting in which any point in the simplex can in principle be sampled and each proxy run is treated as having the same cost. Practical data-mixing studies may face further constraints, including infeasible domain proportions, data availability, licensing, deduplication, filtering requirements, unequal preprocessing costs, and other operational restrictions. In applied pretraining studies, these restrictions should be incorporated into the design problem itself, either by constraining the feasible mixture region or by using cost-sensitive design strategies \citep{Mohammed2026optimal}.

Future work can extend the present framework in several directions. First, model scale and training conditions can be incorporated directly into the response-surface model as process variables. In the present study, the response surface is fitted at the proxy scale and then evaluated by its ability to preserve rankings at larger scales. A more integrated formulation would model scale transfer itself by treating model size or token budget
as process variables interacting with the mixture proportions. This corresponds to a mixture-process experiment: the mixture variables determine the relative composition of the training data, while the process variables determine the conditions under which that mixture is used \citep{piepel1985}. This idea is standard in classical mixture experimentation. For example, in industrial formulation studies, the final response may depend not only on ingredient proportions but also on processing conditions such as temperature, pressure, or mixing time; mixture-process models therefore include both mixture terms and mixture-by-process interactions \citep{Cornell1988Analyzing}. In the present context, the analogous question is whether the effect of a data mixture changes with model scale or training regime.

Second, future work could develop sequential or adaptive designs for data-mixing experiments \citep{Chernoff1959Sequential}. The designs considered in this paper are fixed before proxy training begins. This is appropriate for comparing design geometries, but it may not be the most efficient strategy when proxy runs are expensive. A sequential design could use early proxy results to update the response-surface model, refine uncertainty about important interaction terms, and allocate later runs to regions of the simplex that are most informative for ranking or optimisation. Such a strategy would be especially useful in high-dimensional data-mixing problems, where the number of possible domain combinations grows quickly and a one-shot design may spend resources on regions that are uninformative or practically unattractive.

Third, the framework is not limited to language-model pretraining. Any data-mixing problem with a fixed training budget allocated across data sources has the same simplex geometry. 
This includes multimodal training where image, text, video, and audio data compete for a fixed budget \citep{aghajanyan2023scaling}; domain-adaptive training where general and domain-specific corpora must be balanced \citep{gururangan2020dont}; multilingual training where tokens are allocated across languages or regions \citep{conneau2020unsupervised}; and supervised or instruction-tuning settings where examples from different task families are mixed \citep{longpre2023flan}.
In all these cases, the key experimental unit is not only a model or a dataset, but a deliberately chosen allocation of training budget across data sources. The broader implication is that data mixing should be treated as an experimental design problem whenever the composition of the training data is a controllable part of the learning system.

\section*{Acknowledgements}

We thank Rob Deardon for his helpful comments and suggestions on this manuscript.

\section*{Data and Code Availability}
The data and code used in this study will be made publicly available in a GitHub repository upon publication of the article.

\bibliography{ref}

\end{document}